\documentclass[conference]{IEEEtran}
\IEEEoverridecommandlockouts
\usepackage{cite}
\usepackage{amsmath,amssymb,amsfonts}
\usepackage{algorithmic}
\usepackage{graphicx}
\usepackage{textcomp}
\usepackage{xcolor}
\def\BibTeX{{\rm B\kern-.05em{\sc i\kern-.025em b}\kern-.08em
    T\kern-.1667em\lower.7ex\hbox{E}\kern-.125emX}}
\begin{document}

\title{Encoder-Decoder Networks for Self-Supervised Pretraining and Downstream Signal Bandwidth Regression on Digital Antenna Arrays\\
\thanks{This work was funded under US Defense Advanced Research Projects Agency agreement HR00112190100.}
}

\author{
\IEEEauthorblockN{Rajib Bhattacharjea}
\IEEEauthorblockA{
\textit{DeepSig, Inc.}\\
Atlanta, Georgia, USA \\
raj@deepsig.io}
\and
\IEEEauthorblockN{Nathan West}
\IEEEauthorblockA{
\textit{DeepSig, Inc.}\\
Rosslyn, Virginia, USA \\
nwest@deepsig.io}
% \and
% \and
% \and
% \and
}

\maketitle

\begin{abstract}
    This work presents the first applications of self-supervised learning applied to data from digital antenna arrays. Encoder-decoder networks are pretrained on digital array data to perform a
    self-supervised noisy-reconstruction task called channel in-painting, in which the network infers the contents of array data that has been masked with zeros. The self-supervised step requires
    no human-labeled data. The encoder architecture and weights from pretraining are then transferred to a new network with a task-specific decoder, and the new network is trained on a small
    volume of labeled data. We show that pretraining on the unlabeled data allows the new network to perform the task of bandwidth regression on the digital array data better than an equivalent
    network that is trained on the same labeled data from random initialization.
\end{abstract}


%%%%%%%%%%%%%%%%%%%%%%%%%%%%%%%%%%%%%%%%%%%%%%%%%%%%%%%%%%%%%%%%%%%%%%%%%%%%%%%%
\section{Introduction}
Digital antenna arrays produce volumes of radio frequency (RF) data that are often too large to transfer over a single standard high-speed interface such as 100GbE or PCIe. This is because a typical 
digital array samples waveforms at each antenna element at tens or hundreds of megasamples per second (e.g., using an integrated processor like the Xilinx RFSoC as in \cite{Kolodziej8702}),
with each sample typically having 24-28 bits of precision and requiring 32 bits to transfer.
Furthermore, the total data rate out of an array scales linearly with the desired instantaneous bandwidth and with the number of antenna
elements, meaning that a wideband array rapidly saturates a high-speed interface and computational resources as the bandwidth and number of elements increases. The main method currently used to
address this issue is to form a weighted sum of several antenna element signals into a single signal, which is known as beamforming and effectively reduces the data rate and increases the system
sensitivity to signals impinging from certain directions while attenuating signal sensitivity from other directions. Note, however, that this operation loses information about signals from some
directions. There are, however, many degrees of sparsity/redundancy/structure in the signals that could be exploited to better reduce the volume of data coming out of a digital array. For example,
the received signal spectrum at each element of an array is very similar; each element simply has a position-dependent spectral amplitude and phase offset relative
to other elements in the array. These offsets are themselves a relatively low-information / highly structured function of space and frequency, and can be captured by a few coefficients in a
spatial Fourier representation, with each coefficient corresponding to a multipath direction-of-arrival. Similarly, in a typical terrestrial environment, only some fraction of the spectrum is
occupied, which means that, under the current approach, ADC samples are being used to represent a lot of noise that exists between actual signals. Finally, the types of signals that 
need to be represented themselves have structure that distinguishes them from random noise, which means that encodings of these data exist that can store them in fewer bits than their original
format that comes out of the array. Current approaches to array signal processing do not leverage this spatial redundancy, spectral sparsity, and temporal structure, meaning that the amount of data
coming out of the array is vastly larger than what is strictly necessary to represent the signals.

Data-driven machine learning has the potential to address this issue, particularly through the application of self-supervised learning (SSL). In its current incarnation, SSL uses neural
network models to learn compressed representations of data distributions. These compressed representations are known as embeddings, and have far fewer degrees of freedom and far less
dimensionality than the original data sources. Because digital array data contains redundancies and sparsity, it is highly compressible without loss of information, and it is expected that
learned compression methods using SSL will discover how to leverage this redundancy and sparsity for compression, making digital array data an especially good candidate for the SSL approach.
Embeddings are learned from input data that requires no human labeling, which contrasts with the current wave of supervised methods in deep learning.
Requiring large volumes of labeled data has been identified as one of the factors that makes machine learning less attractive in real RF applications \cite{Wong2022}, which is another factor
that makes SSL an ideal candidate for further explorations with RF data.
A network that generates embeddings is derived through a pretraining process on unlabeled data. One approach to pretraining is to solve a related problem that does not require
human generated labels, known as a pretext task. For example, an encoder-decoder network can be trained to undo an input transformation such as the addition of noise or zeroing of some
input data. At the completion of such a pretraining process, the encoder network is taken as the embedding generator for further downstream tasks. The pretrained network (encoder) outputs
(the embeddings) are then used as inputs for further downstream machine learning based algorithms. In the case of digital antenna arrays, downstream algorithms may include beamforming weight
estimation, signal detection in noise, or joint signal detection and direction-of-arrival estimation.

To summarize, the potential benefits of an SSL-based approach to machine learning on digital antenna arrays are twofold: 1) embeddings of the array data can be transferred in place of
the raw data over a high-speed interface out of the array, and 2) far less labelled data can be used to train downstream machine learning algorithms performing functions of array signal 
processing interest. To further try to realize these benefits, we present in this paper the first investigations into self-supervised learning on antenna array data. Our specific contributions include:
\begin{itemize}
\item A new proposal for a pretext task for the RF array domain, and
\item The first known demonstration showing that pretraining on a pretext task and transferring on a small amount of labeled data to a downstream RF task on an array produces superior results to
training the same downstream model from random initialization.
\end{itemize}

The rest of the paper is organized as follows. Section \ref{sec:related_work} provides an overview the developments in machine learning that are related to the present work. Section
\ref{sec:methods} describes the data collection, hardware, and machine learning methodologies used in our applications of both self-supervised pretraining and downstream task training.
Section \ref{sec:results} demonstrates some sample experimental results for both pretraining and downstream task training. Section \ref{sec:discussion} concludes with a discussion of the
results, limitations of the present study, and future directions for research.
%%%%%%%%%%%%%%%%%%%%%%%%%%%%%%%%%%%%%%%%%%%%%%%%%%%%%%%%%%%%%%%%%%%%%%%%%%%%%%%%
\section{Related Work}\label{sec:related_work}
This work builds on convolutional neural networks and on self-supervised representation learning from the computer vision and natural language processing domains, and so we briefly provide an
overview in this section of those areas of machine learning.
%%%%%%%%%%%%%%%%%%%%%%%%%%%%%%%%%%%%%%%%%%%%%%%%%%%%%%%%%%%%%%%%%%%%%%%%%%%%%%%%
\subsection{Supervised Convolutional Neural Networks}
Data-driven machine learning using neural-network models has undergone a renaissance since AlexNet\cite{Krizhevsky2012} was submitted to the 2012 ImageNet Large Scale Visual
Recognition Challenge (ILSVRC)\cite{Deng2009}.
AlexNet leaped past the 2010 and 2011 winners by 12 and 10 percentage points of accuracy, respectively.
Neural networks dominated the ILSVRC competition through its conclusion in 2017, and were also rapidly adapted to other tasks
in other problem domains such as generative image modeling\cite{Goodfellow2014}, audio classification\cite{Oord1807}, speech synthesis\cite{Oord1609}, speech recognition\cite{Hannun2014},
and generative text modeling\cite{Dauphin2016}.
These methods were first applied to the radio frequency (RF) signal domain in 2016 for modulation format recognition \cite{OShea2016} using data from a single simulated antenna, which
lead to other works in the modulation recognition problem using similar datasets\cite{Krzyston2020}. The current work is in this lineage of applications of convolutional networks to problems in
the RF domain.
%%%%%%%%%%%%%%%%%%%%%%%%%%%%%%%%%%%%%%%%%%%%%%%%%%%%%%%%%%%%%%%%%%%%%%%%%%%%%%%%
\subsection{Self-Supervised Learning}
The area of self-supervised pretraining of neural networks was similarly revived in the post-AlexNet era, with major advances first in natural language modeling \cite{Mikolov2013},
followed by demonstrations of self-supervised pretraining in the image domain\cite{Zhang2016,Doersch2015} and speech recognition domains\cite{Synnaeve2016}, among others. The application of
self-supervised pretraining on RF data is rare in the literature, with typical examples involving radar return data that has been processed into an image format \cite{Alloulah2021,Li2019}.
In the case of \cite{Alloulah2021}, the self-supervision takes the form of making embeddings of radar return data consistent with corresponding pretrained image embeddings derived from a
conventional camera input. Reference \cite{Li2019} trains a network to approximate a conventional radar motion tracking algorithm, which, while not requiring human labels, can only learn to approximate a human expert algorithm.
Notably, there is a gap in the literature in the area of self-supervision on RF data directly from antenna arrays.
%%%%%%%%%%%%%%%%%%%%%%%%%%%%%%%%%%%%%%%%%%%%%%%%%%%%%%%%%%%%%%%%%%%%%%%%%%%%%%%%
\section{Methods}\label{sec:methods}
%%%%%%%%%%%%%%%%%%%%%%%%%%%%%%%%%%%%%%%%%%%%%%%%%%%%%%%%%%%%%%%%%%%%%%%%%%%%%%%%
\subsection{Data Collection Hardware and Dataset Creation}
\begin{figure}
    \centering
    \includegraphics[width=3.4in]{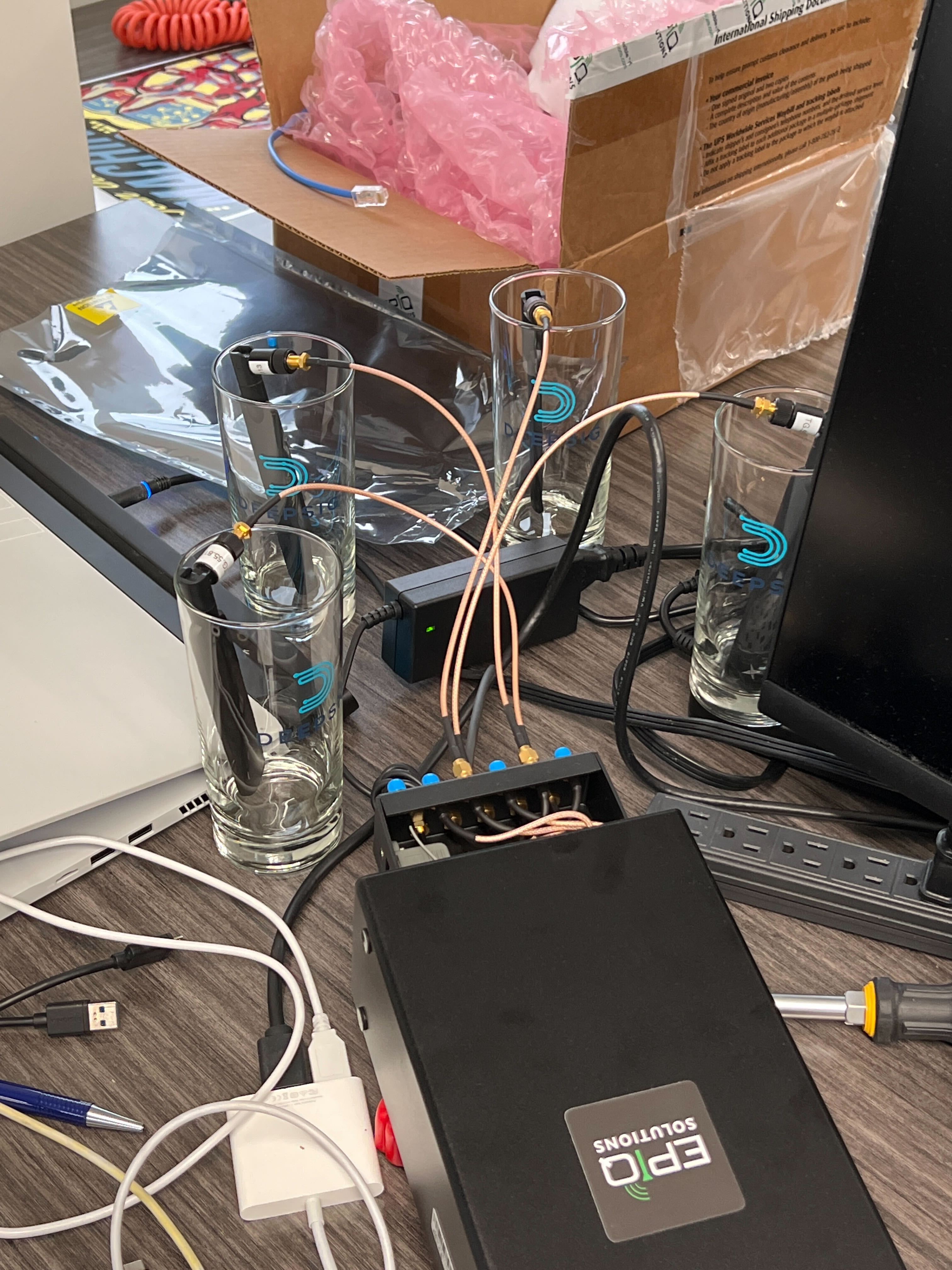}
    \caption[An irregular array geometry is depicted.]{An irregular array geometry of convenience was created using four Taoglas Limited TG.55.8113 antennas and DeepSig Inc. drinking glasses as a
    physical support. Our machine-learning based approach requires no explicit calibration of phase and no explicit representation of the array response, so such irregular geometries pose no distinct
    problems. This geometry was used to perform RF data collection in Rosslyn, VA in the winter of 2021.
\label{fig:drinking_glasses}}
\end{figure}

An Epiq Solutions Sidekiq X4 software-defined radio (SDR) with 4 channels has been used for data collections. This radio downconverts and digitizes 4 channels
at up to 250 MS/s. For this effort, it has been operated at 50 MS/s with an analog channel bandwidth of 41 MHz after accounting for filter roll-off effects.
The physical aperture configuration was driven by simplicity and cost: simple monopole antennas (Taoglas Limited TG.55.8113) have been put into support structures of convenience, leading to irregular
array geometries without a specific spacing pattern. This reduces costs by not requiring the construction of any additional physical structure to support the antennas. Being able to use arbitrary
antenna geometries is enabled by the fact that the machine learning algorithms learn the array response and characteristics implicitly without any explicit calibration. The irregular array 
used for data collection is displayed in Figure \ref{fig:drinking_glasses} and was used to collect a four-channel dataset from the Epiq Sidekiq X4.
The Epiq Sidekiq X4 was tuned to 115 different frequencies starting at 375 MHz and ending at 2573 MHz.
Each center frequency for recording was chosen by a human operator looking at a spectrogram visualization and trying to minimize the presence of signals crossing the upper and lower band edges.
Recordings were collected at each frequency for 100 ms, which corresponds to 5 million samples at the 50 MS/s rate. The dataset is read into the machine learning framework PyTorch. The entire dataset
can be thought of as a large tensor with shape $[115, 4, 5000000, 2]$: 115 frequencies, 4 antenna channels, 5000000 time steps, and 2 quadrature channels.
%%%%%%%%%%%%%%%%%%%%%%%%%%%%%%%%%%%%%%%%%%%%%%%%%%%%%%%%%%%%%%%%%%%%%%%%%%%%%%%%
\subsection{Pretraining Methods}\label{sec:pretraining}
\begin{figure}
    \centering
    \includegraphics[width=3.4in]{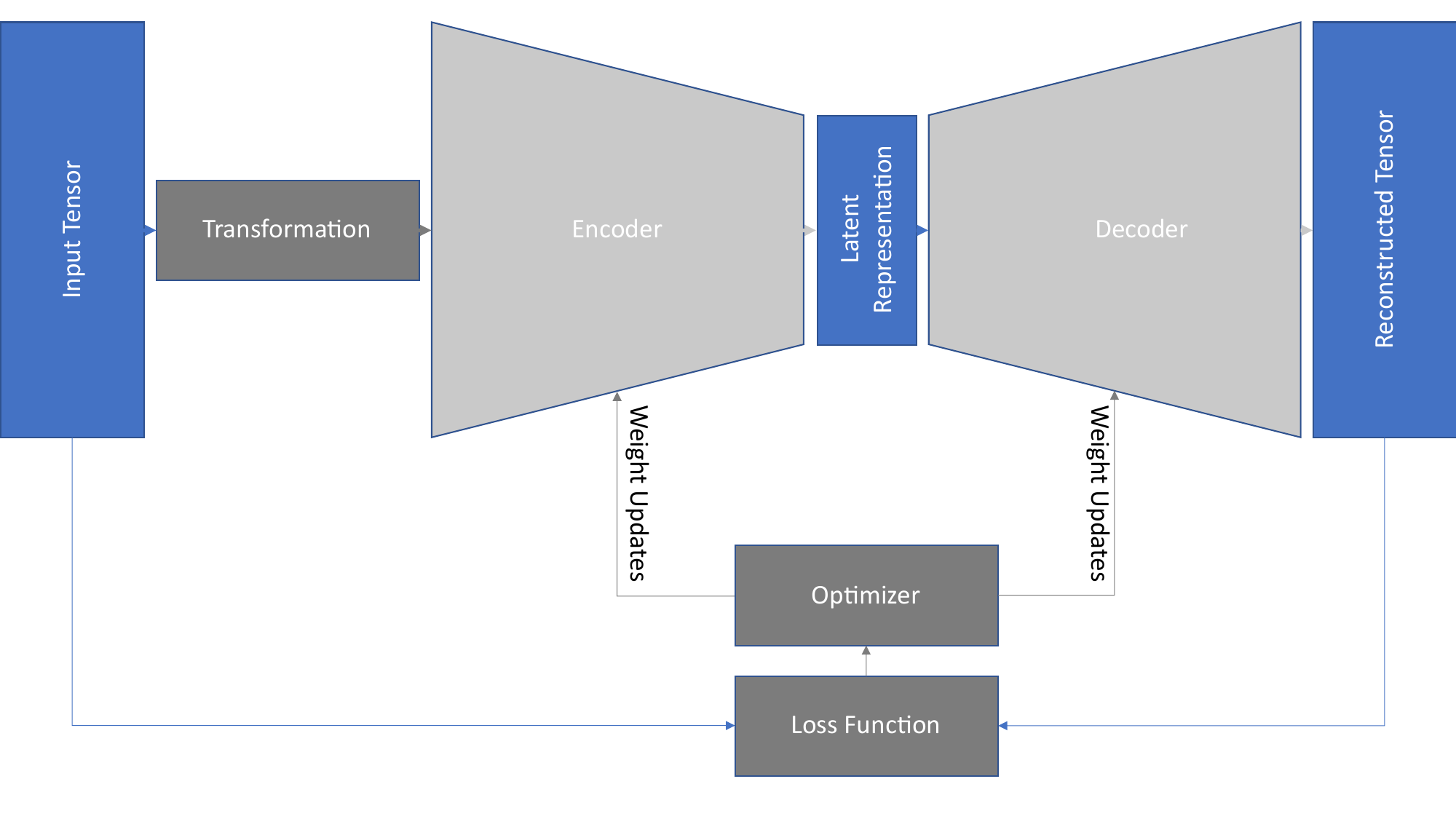}
    \caption[Encoder-decoder architecture for the reconstruction task.]{The process for pretraining an encoder-decoder network for learning embeddings is depicted. Data samples from the input
    distribution (e.g., file captures from a multichannel SDR) are passed through a transformation that corrupts the data, seen on the left side of the figure.
    The network's pretext task is to reconstruct the original data from the corrupted data. This pretext task does not require any ground-truth labels related to the input tensor. 
    The encoder neural network reduces the dimensions of the data representation, arriving at a latent representation that serves as the learned embedding, seen in the center of the figure.
    The latent representation is passed through a decoder neural network, which outputs a tensor with the same shape as the input tensor, seen in the right half of the figure. The input tensor and
    the output tensor are compared using a loss function (e.g., mean-squared error between the tensors), and the loss function is minimized by an optimizer, both seen at the bottom of the figure.
    Once the loss has been minimized, the encoder has learned a mapping from corrupted input samples to a latent representation that contains enough information to reconstruct the uncorrupted
    data to some level of fidelity, but with far less dimensionality.
\label{fig:encoder_decoder_arch}}
\end{figure}

We have focused on convolutional encoder-decoder networks trained on reconstruction pretext tasks on spectrogram data from an array. The pretraining
step is depicted in Figure \ref{fig:encoder_decoder_arch}.
Pretraining proceeds by minimizing an error measure between the encoder-decoder network output and the original data. This is a reconstruction pretext
task, and the particular transformation we will present results for in this document is a channel-masking operation called channel in-painting that uniformly-randomly selects the data for one
antenna channel per training
example and sets that data to all zeros. If the network succeeds in reconstructing the original data from this transformed data, then the network has performed ``channel in-painting'', which
also refers to similar pretext tasks from natural language processing and computer vision. The remainder of this section describes the implementation details of how we pretrained a
network to perform channel in-painting on digital array data.

The data tensors into this network are derived from frames RF data from the dataset. Examples are drawn from the dataset with a shape of $[4,65536,2]$, corresponding to $4$ antenna channels, $65536$
time samples, and
$2$ quadrature channels. These are frames of raw time-domain input from the array over $1.311$ ms of continuous time. These data are pre-processed into a joint time-frequency representation
(short-time Fourier transform) suitable
for detection of signals in noise. First, the trailing dimension of 2 on the tensors is absorbed into the time dimension by converting the tensor datatype to be complex.
This results in a tensor with shape $[4, 65536]$. The time dimension is reshaped into two dimensions with $32$ time chunks and $2048$ continuous time steps, resulting in a tensor of shape
$[4, 32, 2048]$. After applying a Hann window function along the time dimension of length $2048$, a discrete Fourier transform is calculated along that dimension, resulting in $4$-channel
Hann-windowed
short-time Fourier transforms (STFTs) with 32 time chunks and 2048 frequency bins (a shape $[4, 32, 2048]$ complex tensor). The real and imaginary parts of the tensor are now separated back
into a new trailing dimension with size 2, giving a tensor of shape $[4, 32, 2048, 2]$ with all real entries. The trailing quadrature dimension is merged with the antenna array channels,
resulting in 8-channel training examples, with each channel containing the real or imaginary component of the STFT of the signals at the antennas (shape $[8, 32, 2048]$).
Finally, each training example is normalized (standardized) by subtracting its mean value and dividing by its standard deviation, giving each training example zero mean and unit
variance across the channels, time, and frequency dimensions.

\begin{figure}
    \centering
    \includegraphics[width=3.4in]{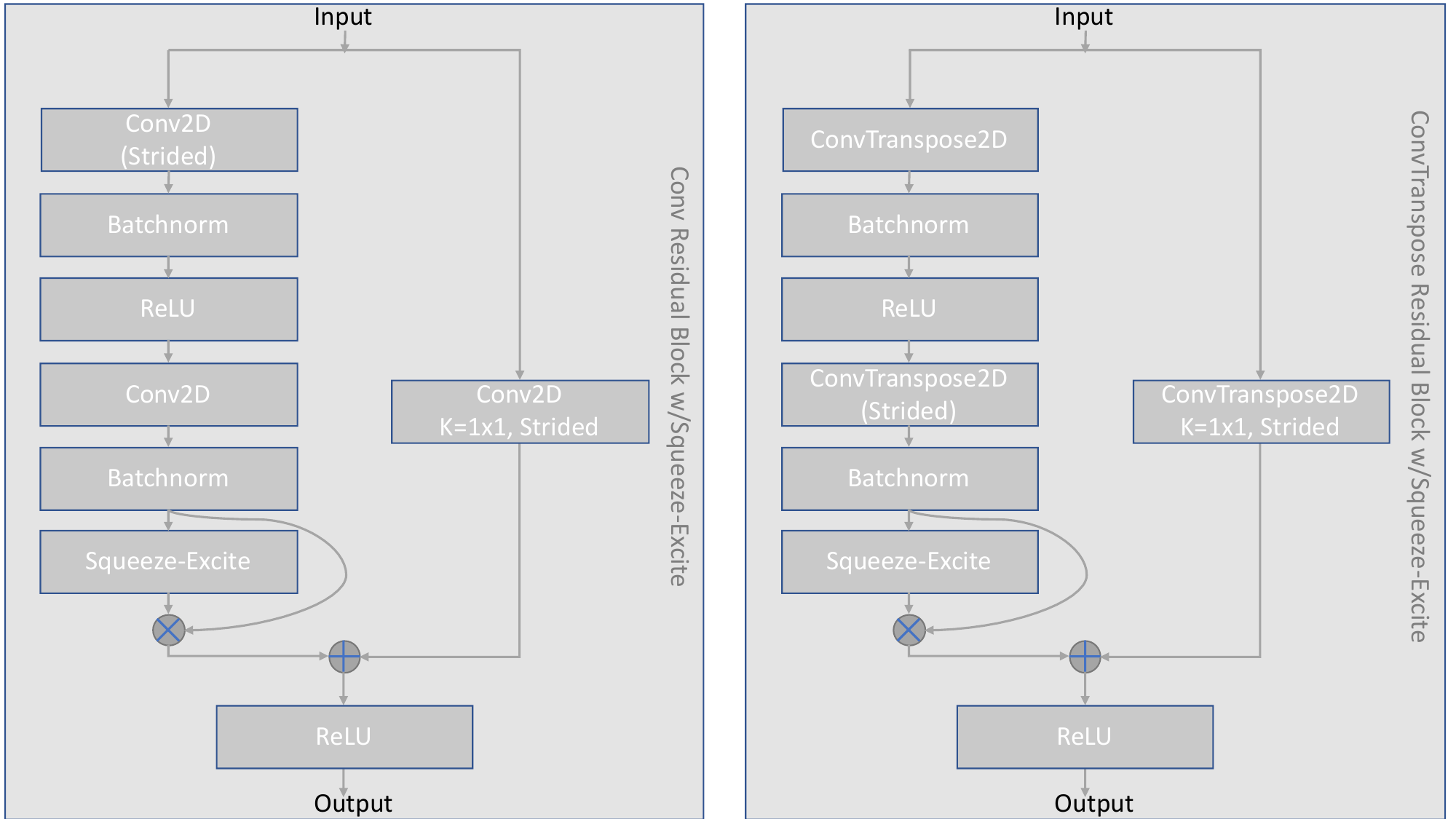}
    \caption[The convolutional building blocks of the network.]{The two variants of the convolutional residual blocks with squeeze-and-excitation are depicted. The left block reduces
    the dimension of the data (downsampling) using strided convolution. The residual path has a 1x1 convolution used for matching the downsampled output dimensions of the other path.
    The right block is nearly identical, but it increases the dimensions of the data (upsampling) using strided transposed-convolution. These blocks are parameterized by 2D kernel size, 2D stride,
    input channel count, and output channel count.
\label{fig:block_architecture}}
\end{figure}

The encoder-decoder architecture used is based on basic principles of reducing the dimensionality of the data and then increasing it again. The core of the network is based on a common
convolutional residual block structure with squeeze-and-excitation with a squeeze reduction ratio of eight \cite{Hu2017}. There are two variants of this
common structure: one that reduces dimensionality and one that increases dimensionality. The two variants of this basic building block are depicted in Figure \ref{fig:block_architecture}.

\begin{figure}
    \centering
    \includegraphics[width=3.4in]{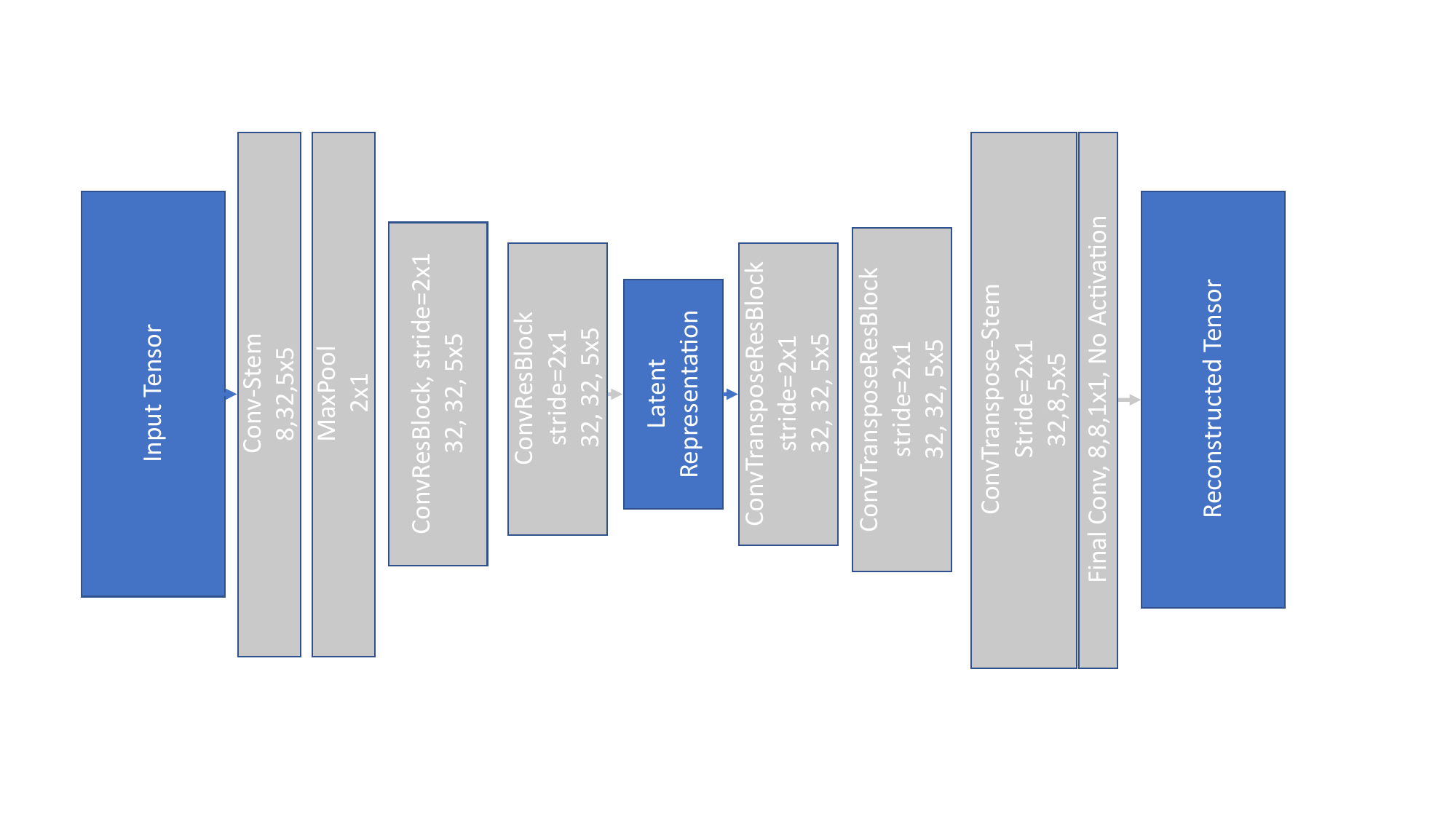}
    \caption[An exemplar encoder-decoder network.]{An example encoder/decoder structure for 4-channel antenna data is presented. The data has 8 input channels
    each of which is the real or imaginary part of a windowed STFT of an original time-domain sequence with length 65536. A convolutional stem increases this channel count to 32 channels
    while keeping the time and frequency resolution constant. Next a pooling layer reduces the STFT time resolution by 2. Two subsequent layers downsample using strided convolution,
    arriving at the latent representation of the data with 32 channels, 4 STFT time steps, and 2048 frequency bins. The decoder operates in the transposed fashion such that strided layers
    perform an upsampling operation in the STFT space. All layers are specified by their 2D stride (explicitly specified by the string ``stride='', with a default of stride=1 assumed if
    not specified), and a weight tensor shape, written in the format (in\_channels, out\_channels, kernel\_size), where the kernel size is written with an ``x'' between the two size dimensions
    (read as the word ``by'', e.g., 5x5 is a 5-by-5 kernel size).
\label{fig:exemplar_architecture}}
\end{figure}

Many network architectures based on stacking this block have been explored. One representative architecture is presented in Figure \ref{fig:exemplar_architecture}.
The architecture initially increases the channel count of the data from 8 to 32 using a convolutional stem (a simple block consisting of a single convolutional layer, batch normalization,
ReLU activation, and squeeze-excite), which increases the data tensor sizes by 4x. It then proceeds to decimate by two in time while keeping the channel count and number of frequency bins constant
a total of three times, representing an 8x decrease in the data tensor size. In total, the latent representation is half the size of the input tensor.
The network presented in Figure \ref{fig:exemplar_architecture} will be referred to as the ``2-resblock'' network, indicating that it uses two serial downsampling resblocks in the encoder and two
serial upsampling resblocks in the decoder. Other designs with more resblocks were also explored.

All networks were pretrained by a common procedure based on the Adam variant of stochastic gradient descent using mini-batches containing 16 examples each (mini-batch shape $[16, 8, 32, 2048]$)
and an initial learning rate of 0.001. Each mini-batch was drawn from the transformed data in which 2 adjacent channels out of the 8 were randomly set to zero for a channel in-painting task.
The loss metric was the mean-squared-error over the mini-batch, and one gradient-based update was performed per training mini-batch using the Adam optimizer. Twenty percent of the training data was
withheld for validation (7084 training examples and 1771 withheld validation examples). A total of 443 mini-batches are processed through the network each epoch, with one gradient-based weight update
per mini-batch. The average loss over the mini-batches was recorded for each epoch. The average loss over the validation data is calculated and recorded at the end of each training epoch. If the
validation loss is lower than any previous validation loss, the network is serialized and saved to disk as one of the outputs of the training procedure. The number of epochs the network is trained
for is determined by validation-based early stopping with a patience-parameter of 30, meaning that if the validation loss does not decrease once in any 30-epoch window, training stops. Finally, the learning
rate is also adapted throughout training using the “reduce-on-plateau” method with a patience-parameter of 10, meaning that if the validation loss does not decrease once in any 10 epoch window, then
the learning rate is reduced by a factor of 10.
%%%%%%%%%%%%%%%%%%%%%%%%%%%%%%%%%%%%%%%%%%%%%%%%%%%%%%%%%%%%%%%%%%%%%%%%%%%%%%%%
\subsection{Downstream Task Definition and Methods}
\begin{figure}
    \centering
    \includegraphics[width=3.37in]{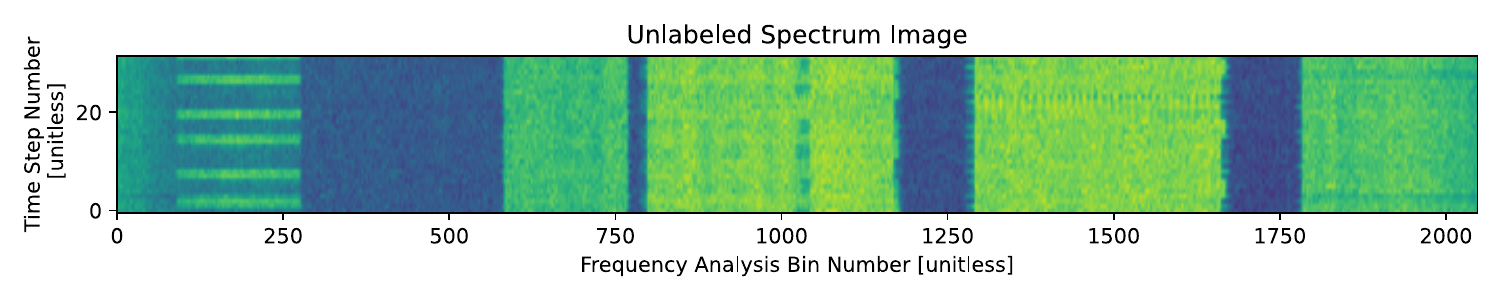}
    \includegraphics[width=3.4in]{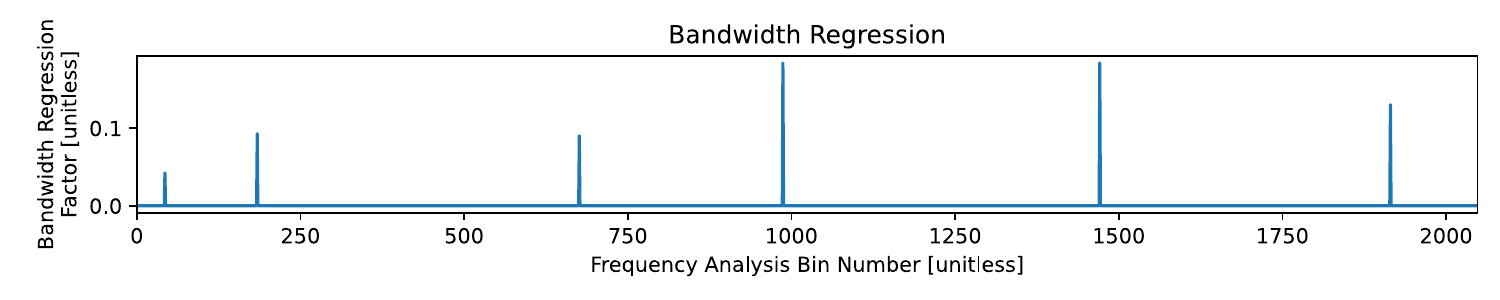}
    \caption[An example pair of spectral representation and bandwidth regression target.]{The spectrogram for a spectrum capture is depicted in the top panel. The picture results from
    taking a chunk of 65536 RF samples from one antenna and calculating the spectrogram (the log-magnitude of the Hann-windowed and non-overlapping STFT on 2048 analysis bins). Six signals
    can be seen across the frequency bins, even if only partially, with varying power levels indicated by the intensity of the colors, and varying bandwidths, indicated by the width
    in frequency analysis bins occupied by the signal. The bottom panel
    depicts the regression target values, which are nonzero only where the signal centers are specified, and take on values proportional to the bandwidth of the corresponding signal. Note there are
    six such target values, one for each signal in the panel above. The
    machine learning task of wideband signal bandwidth regression is mapping multichannel complex STFT data (the log magnitude of a single channel of which appears in the top panel) to the
    bottom sequence.
    Note that the representation above is flexible enough to distinctly represent signals that overlap in time and frequency, so long as their center frequencies differ by the width of
    at least one analysis bin. For this effort, we specifically map from 4 such channels of STFT data to a single estimate of bandwidth regression values.
    The system is flexible and can work on any number of input antenna channels, with only reduction in the real-time processing rate for larger numbers of antenna elements.
\label{fig:bw_regression_example}}
\end{figure}

After pretraining, a new encoder-decoder network is initialized with the same encoder architecture and weights as the pretrained network, but with a randomly initialized task-specific decoder
that outputs a tensor shape for another task, not the channel in-painting pretext task. In this work, we consider the downstream task of signal bandwidth regression, which is the act of mapping
between the STFT of a signal and a function on the frequency bins that takes on a value proportional to the signal bandwidth in each signal center bin and takes a value of zero everywhere else.
Figure \ref{fig:bw_regression_example} depicts an example-pair of a spectrogram derived from STFT data, and the corresponding bandwidth regression target.

The regression target values themselves are simply a normalized version of the signal bandwidth such that a signal that occupies all frequency bins gets a target value of 1 ($1/2048$ times the number
of bins the signal occupies).
To map from the pretrained embeddings to the bandwidth regression targets, the new decoder continues to downsample the embedding in the time dimension until
that dimension has size 1, at which point the singleton dimension is removed, leaving a tensor of shape $[32, 2048]$, corresponding to $32$ different features across $2048$ frequency bins.
To map from this shape to the desired target bandwidth regression shape of $[1, 2048]$, 1D convolutional residual blocks are used with no upsampling or downsampling. These have a similar structure
to those depicted in Figure \ref{fig:block_architecture}, except all convolutions are replaced with 1D convolutions, all strides are set to one, and the 1D form of squeeze-and-excite is used.
Each block halves the channel count, and so after a stack of 5 such blocks, the decoder output shape matches that of the bandwidth regression target. The kernel size is set to 5 in these 1D
convolutional blocks. The residual path still has size 1 convolutions to match the channel counts between input and output.

The training procedure and hyperparameter choices are identical to those of pretraining (described previously in Section \ref{sec:pretraining}), except the initial
learning rate of Adam was 0.01. The loss function being optimized is a mean-squared error loss defined on the logarithm of the bandwidth targets, with a small
epsilon hyperparameter added to the network output for numerical stability. Let $B_i$ for $i\in\{0,1,\ldots 2047\}$ be the sequence of true bandwidth regression target
outputs on the 2048 analysis bins as depicted in the bottom panel of Figure \ref{fig:bw_regression_example}, and let $\hat{B}_i$ be a sequence that will estimate the true
bandwidth regression target, i.e., the output of the downstream neural network. If we let $B$ and $\hat{B}$ without subscripts represent their entire respective sequences,
then the bandwidth loss between a true and estimated bandwidth regression target is mathematically defined by
\begin{equation}
    \ell(B,\hat{B})=\sum_{\{i|B_i\neq 0\}} \left(\log(B_i)-\log(\varepsilon+B_i)\right)^2
\end{equation}
The optimizer tries to minimize the average of that quantity over the training dataset by training over minibatches. Critically, this loss is only calculated on the bins with non-zero target values. This is
a pure bandwidth regression problem, and effectively assumes that the signal centers are known by some other method to the network. In the most useful general case, the network
should itself regress both the bandwidth regression values, and their frequency locations. That this work does not present those results is a current limitation that will be addressed in Section
\ref{sec:discussion}.

This downstream bandwidth regression problem requires a labeled dataset, and a new single file was recorded that contains 77 training examples. These examples were labeled by upper and lower
signal band edges, which are then processed into the ground-truth bandwidth regression targets. The labeled dataset is itself split into 62 training examples and 15 validation examples. This
is a very low data regime for learning with neural networks, but if the network can generalize to the 15 validation examples by only training on the 62 training examples, it can be concluded
that pretraining was effective in helping the network generalize even with small amounts of data.

\section{Sample Experimental Results}\label{sec:results}
%%%%%%%%%%%%%%%%%%%%%%%%%%%%%%%%%%%%%%%%%%%%%%%%%%%%%%%%%%%%%%%%%%%%%%%%%%%%%%%%
\subsection{Pretraining Results}\label{sec:pretraining_results}
\begin{figure}
    \centering
    \includegraphics[width=3.4in]{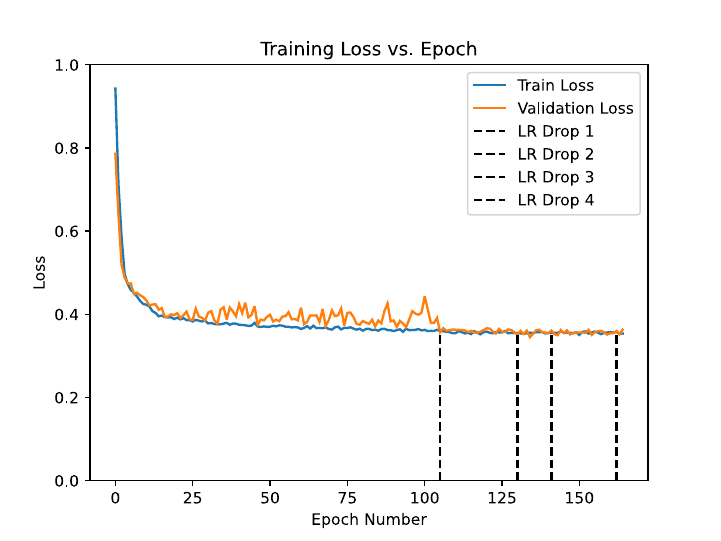}
    \caption[The pretraining loss curves achieved during pretraining for the 2-resblock architecture.]{The training and validation loss of the 2-resblock architecture are presented.
    The model trained for a total of 164 epochs before early stopping after 4 learning rate drops, and achieved a lowest validation loss of 0.3452 on the held-out data examples.
\label{fig:pretrain_loss_2resblock}}
\end{figure}

The 2-resblock variant of the convolutional encoder-decoder network was pretrained on the channel in-painting task using the dataset and training procedure previously outlined in
Section \ref{sec:pretraining}. The training results are presented in Figure \ref{fig:pretrain_loss_2resblock}.
The 2-resblock model trained for a total of 164 epochs, with the model from epoch 135 achieving the lowest validation loss of 0.3452. The mean-squared error was used as the loss measure for
optimization, and the data inputs are zero-mean, unit-variance when those quantities are calculated across all dimensions for a training example. As a reference for the loss scale, two tensors
with entries drawn from independent identical normal distributions with zero mean and unit variance would have an expected mean-squared error loss of 1.0. Because our dataset is standardized
to zero mean and unit variance, that means that a loss value of 1.0 indicates that the network can match the mean and variance of the data, but nothing else. On the other hand, a network that
correctly reconstructs the real and imaginary parts of every time step and frequency bin of all four antenna channels for every training example would achieve a loss of zero. Note that the
training procedure does drive the loss down from a value near 1.0, meaning that the network initially can only match the mean and variance of the data distribution but learns to represent the
data distribution better than that as training proceeds.
%%%%%%%%%%%%%%%%%%%%%%%%%%%%%%%%%%%%%%%%%%%%%%%%%%%%%%%%%%%%%%%%%%%%%%%%%%%%%%%%
\subsection{Downstream Transfer Results}\label{sec:downstream_results}
\begin{figure}
    \centering
    \includegraphics[width=3.4in]{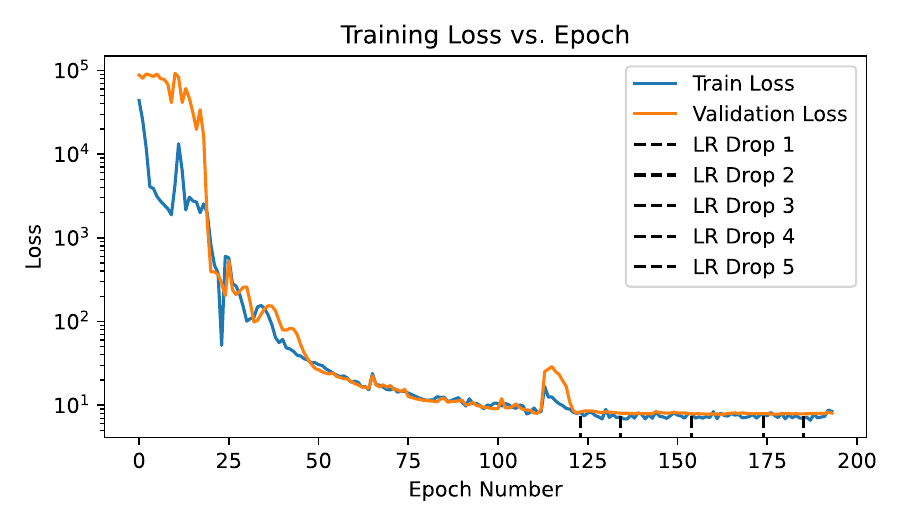}
    \includegraphics[width=3.4in]{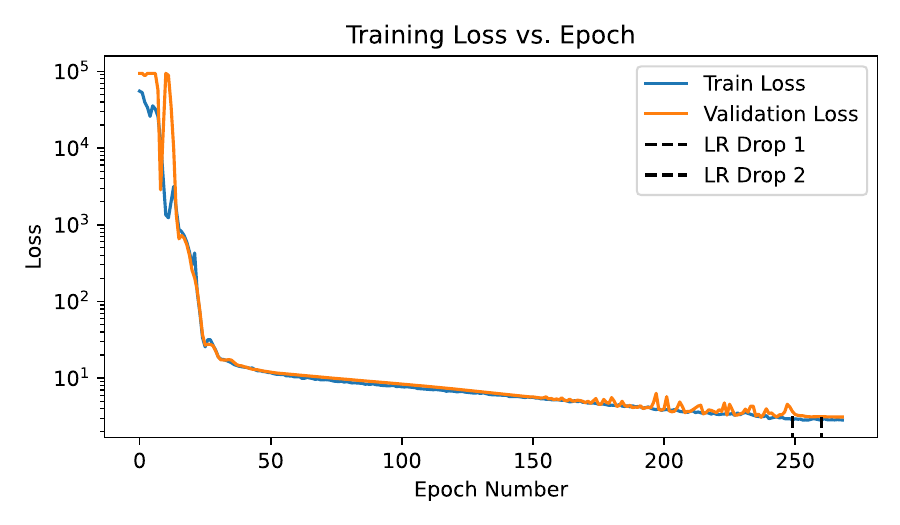}
    \caption[Comparison of the training loss curves for a pretrained and randomly initialized network.]{The top panel depicts the training and validation loss curves for the baseline model with
    random initialization, trained on 62 training examples and evaluated on 15 validation examples. The best mean validation loss in the top panel is 7.7. The bottom panel depicts
    the training and validation loss curves for a model for which the encoder was pretrained on a channel in-painting task over a much larger dataset of more than 7000 training examples.
    The best mean validation loss in the bottom panel is 3.0.
\label{fig:baseline_training_loss}}
\end{figure}

For transfer learning from the pretrained model to the downstream task of bandwidth regression, we present two sets of results for the same encoder-decoder architecture: 1) the baseline has a
randomly initialized encoder and decoder, and 2) the pretrained model has the encoder copied from an SSL pretrained network, with the same random initial decoder weights as the baseline. 
We present results here for the case in which both the encoder and decoder are trainable. The
baseline model trained for 194 epochs before early stopping and achieved a best mean validation loss of 7.7, with a best-case loss value of 1.6. The pretrained model trained for 269 epochs and
achieved a best mean validation loss of 3.0, with a best-case loss value of 0.6. The training loss curves are depicted in Figure \ref{fig:baseline_training_loss}.

%%%%%%%%%%%%%%%%%%%%%%%%%%%%%%%%%%%%%%%%%%%%%%%%%%%%%%%%%%%%%%%%%%%%%%%%%%%%%%%%
\section{Discussion, Limitations, and Future Directions}\label{sec:discussion}
That the pre-trained network converged to a lower loss value on the validation data is the key result. To our knowledge, this is the first set of investigations to
demonstrate 1) a pre-text task for digital array data and 2) the utility of SSL pretraining for digital array problems. The method leads to a reduction in volume of required labeled training data,
which could make data-driven machine learning for RF applications more attractive to practitioners. Furthermore, the 2-resblock architecture was shown to have an embedding size that requires
half the number of bits to store as the original STFT data, hinting towards a future in which array data can be massively compressed by learned algorithms.

A notable limitation of the current effort is that the pretrained encoder was trainable (rather than having its weights frozen) in the results shown above. In our experiments
with a frozen pretrained encoder, we found that the loss values on the downstream task were worse than a randomly initialized, fully-trainable baseline. This indicates that
pretraining helps to put the encoder in a better initial state for downstream training and serves as a kind of weight regularization, but that, at least for the scenario we have presented,
the encoder must be trainable on the downstream tasks in order for the final network performance to be improved over the baseline. If this fact holds as a generality for other pre-text tasks,
downstream tasks, datasets, and network architectures, that would severely limit the utility of the embeddings because they would have to be viewed as a starting point for further refinement
rather than a fixed representation that is useful for a myriad of downstream tasks.

Another limitation of this work is the assumption that the signal center frequencies are known {\em a priori} to the bandwidth regression network. Ideally, the network would itself be able to
regress not just the bandwidth values, but also their locations. Doing this requires careful engineering of a suitable loss function, and in simple experiments conducted as part of this work,
we found that simple loss functions resulted in networks that were unable to predict any signal center bins.

To see if the above limitations hold, future research will be directed towards 1) expanding the domain of pre-text tasks (and pretraining methodologies in general), 2) exploring further downstream
tasks (such the full signal detection in noise problem, beamforming weight estimation, direction-of-arrival estimation, etc.), 3) expanding the datasets used for pretraining, 4) exploring
more network architectures, and 5) careful engineering of loss functions suitable for joint signal detection and bandwidth regression. With robust enough pretraining datasets and the appropriate
methodology, SSL has proven to be an effective way to generate fixed embeddings that are useful for a variety of downstream tasks in natural language processing and computer vision. We have little
reason to suspect that there is fundamentally something different about RF data from digital antenna arrays that makes SSL methods fail to provide the same gains to the RF array domain. We
expect it is simply a matter of discovering the correct combination of the factors above that will lead to rapid gains in the field.
\newpage
\bibliography{refs}
\bibliographystyle{ieeetr}
\vspace{12pt}

\end{document}